# Ontology Based SMS Controller for Smart Phones

Mohammed A. Balubaid[1], Umar Manzoor[2]
[1]Industrial Engineering Department, Engineering Faculty
[2]Faculty of Computing and Information Technology,
King Abdulaziz University,
Jeddah, Saudi Arabia

Bassam Zafar[2], Abdullah Qureshi[3], Numairul Ghani[3]
[2]Faculty of Computing and Information Technology
King Abdulaziz University, Jeddah, Saudi Arabia.
[3]National University of Computer and Emerging Sciences,
Islamabad, Pakistan

*Abstract*—Text analysis includes lexical analysis of the text and has been widely studied and used in diverse applications. In the last decade, researchers have proposed many efficient solutions to analyze / classify large text dataset, however, analysis / classification of short text is still a challenge because 1) the data is very sparse 2) It contains noise words and 3) It is difficult to understand the syntactical structure of the text. Short Messaging Service (SMS) is a text messaging service for mobile/smart phone and this service is frequently used by all mobile users. Because of the popularity of SMS service, marketing companies nowadays are also using this service for direct marketing also known as SMS marketing.In this paper, we have proposed Ontology based SMS Controller which analyze the text message and classify it using ontology aslegitimate or spam. The proposed system has been tested on different scenarios and experimental results shows that the proposed solution is effective both in terms of efficiency and time.

I. INTRODUCTION

Mobile phones were initially developed to make and receive calls while being mobile using radio link. Later on, services like text messaging (SMS), multimedia messaging (MMS) were added in the mobile phone devices. In the last two decades, mobile phones have evolved and have become smarter / intelligent devices commonly known as smart phones [1, 2]. Smart phones are built on mobile computing platform and usually have advanced computing abilities / connectivity as compared to the simple mobile phones [8]. Initially smart phones were developed with the integration of mobile phone and personal digital assistant (PDA) functions. These smart phones include number of exciting features like touch screen, mobile web browser (i.e. access websites on mobile phone) and WiFi (i.e. access internet using wireless connection) support.

In year 2000, high resolution touch screen smart phone named Ericsson R380 was released which has its own Operating System. This was first ever smart phone with its own OS, the Operating System used was Symbian OS. In 2005, Google entered into the mobile market with the help of an open source operating system for smart phones called Android. In 2007, Apple [9] introduced a smart phone named iPhone [10] which made big change in the history of smart phones development; Apple development their own Mobile Operating System named as IOS for iPhone and this OS is not open source. Therefore, Android operating system is supported by most of the smart phones companies (such as HTC, Samsung, Sony Ericson).

Along with the launch of iPhone, Apple introduced AppStore (Application Store) where 3[rd] party applications were hosted for distribution (i.e. single platform distribution). Before, Apple AppStore, smart phone applications distribution were largely dependent on third-party sources that developed the application(s) such as GetJar, Handmark, Handango, PocketGear, etc. Application development for Android OS is greatly increasing as compared to IOS because 1) development toolkit is free, 2) Android is open-source, therefore it's easy to integrate applications and 3) Android software suite allows easy integration with Google applications such as Maps, Calendar, web browser etc. Android based smart phones are giving great competition to iPhone.

Text analysis includes lexical analysis of the text and has been widely studied and used in diverse applications. In the last decade, researchers have proposed many efficient solutions to analyze / classify large text dataset, however, analysis / classification of short text is still a challenge because 1) the data is very sparse 2) It contains noise words and 3) It is difficult to understand the syntactical structure of the text [21, 22, 25, 28]. The concept of Short Messaging Service (SMS) was developed in the Franco-German GSM cooperation in 1984 by Bernard Ghillebaert and Friedhelm Hillebrand [11]. SMS is a text messaging service on the phone, web or mobile system and mostly used data application is SMS text messaging. SMS nowadays is also used for direct marketing also known as SMS marketing.

In the last few years, many SMS managers have been developed for managing the SMS on smart phones and the most of them focuses on Spam filtering, Scheduled SMS and automatic-Reply generation. Few popular android applications are 1) Anti SMS Spam: It is a spam filtering application and spams all incoming SMS from unknown numbers when Spam filtering is turned on. 2) Schedule SMS: It is scheduled SMS application and gives time, date, recipient number and text (SMS content) option to the user. The application sends the SMS to the recipient on specified time and date specified by the user. 3) SMS Auto Reply: It is an Auto Reply SMS application which sends an automated reply to all the incoming texts when auto reply is turned on. The content (text) of auto reply is selected / configured by the user.

Ontologies have been widely used for knowledge representation / sharing and have been used in diverse areas [23, 24, 26, 27]. Ontology based SMS Controller is an Android Based Application developed on Android Jelly Beans 4.1, the proposed solution is all in one SMS manager and includes some previous features with advancements as well as some new and exciting Features like ontology based SMS spam detection, Group chat etc. The default android messaging application gives few options to user such as send message / receive message / save message etc. whereas the proposed application provides some additional features in addition the default features. The major features of the proposed application are:

- Automated text replies to messages when a profile is activated
- Scheduled SMS sent on specific dates and events
- Group chat including multiple users like we do in different messengers
- Content based Spam filtering

The above features make the proposed application unique as these features are missing in the existing applications. The remainder of this paper is organized as follows. In Section 2, we present brief overview of related work, this section is followed by the discussion of the Ontology based SMS Controller architecture including the SMS text analysis and classification method. In Section 4, the simulation and experimental analysis of proposed solution is presented. Finally, the conclusion is drawn in Section 5.

## II. RELATED WORK

With the evolution in Smartphone era, leading IT companies and researchers have proposed many efficient applications for the same. In this section, we will review few related applications developed for managing SMS on android platform.

*1) Anti-Spam SMS and Private Box by Droid Mate [12] is an android based anti-spam application with a private box. Its spam feature helps filter unwanted messages from any sender. Key features of this application are: a) Can block SMS from unknown numbers, and b) User can create a block list and can add existing contact or new numbers in the block list.*

*2) Handcent SMS by Handcent Market [13] is an android based SMS scheduling application with the following key features: a) It helps schedule SMS/MMS messages at specific times or at regular intervals e.g. daily and b) It supports blacklist (i.e. deletes incoming SMS / MMS from number in blacklist which also helps block spam messages.*

*3) SMS Scheduling applications: The best three scheduled SMS android applications are as follows: a) Schedule SMS Wishes enables the user to schedule SMS on the contact list (i.e. the user can select the date / time on which the SMS needs to be send to the selected contact or can use the repeat option to send the SMS regularly [daily / weekly] to the selected contacts) b) Google Voice SMS tool enables the user to schedule (i.e. daily, weekly, etc) the SMS with Google voice c) Scheduled Message application enables the user to schedule SMS or email at any given date / time.*

*4) SMS auto reply by Kirill kruchinkin [14] is an auto reply application which sends reply to each incoming SMS when the auto reply option is enabled. The user has to configure the reply to be send when the auto reply option mode is turn on.*

*5) Intelligent auto reply by John Tsau [15] is a rule based SMS application that has Auto Reply and Auto Forwarding features. It automatically replies to the SMS and missed calls according to the rules set by the user.*

*6) GO SMS PRO by GO Dev Team [16] supports the features of scheduled SMS and Group Texting.*

The previous applications include most of the exciting features but they have the following limitations:

- What if a user does not want to spam all the SMS from unknown numbers?
- What if a user wants to spam the SMS from some unusual numbers only?
- What if a user wants his application to auto reply to a certain Group?
- What if a user wants to send different replies to different group of recipients?
- What if a user wants to have all these features in one application?

Ontology based SMS Controller has Solutions to all these questions. It auto reply to a certain Group and sends different replies to different group of recipients. It gives solution for detecting spam SMS using content analysis. Above all these features are all integrated in one application so that a user can easily manage all the features from one application. Plus it includes new features like Group chat and Auto scheduled SMS by synchronizing the events in the Calendar. The key features of the proposed application include:

- Auto-Reply Modes
    - Profile Based,
    - Group Based,
    - Group-Profile Based,
- Auto-Messaging Modes
    - Event mode
    - Birthday mode
    - Scheduled Texts
- Group Chat
- Content based SMS Spam detection
- Auto-Message Report.

## III. SYSTEM ARCHITECTURE

Ontology based SMS Controller is an Android Based Application developed on Android Jelly Beans4.1 as shown in figure 1 and has the following four modules:

- SMS Spam
- Group Chat
- Auto Reply
- Event-Based Messages

Finally, complete content and organizational editing before formatting. Please take note of the following items when proofreading spelling and grammar:

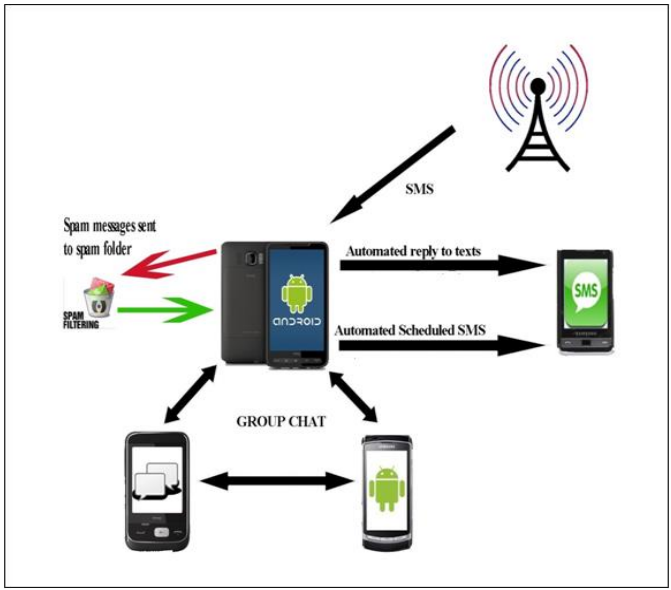

Fig. 1. System Architecture of Ontology based SMS Controller

### A. SMS Spam

For each incoming SMS, the Ontology based SMS Controller interrupts the SMS and executes the SMS Spam classification algorithm to verify the spam messages as shown in figure 2. The SMS Spam classification algorithm is comprised of three steps: (I) Pre-processing (II) Content Analysis (III) Spam Classification.

- **Pre-processing:** In the first step, for each incoming SMS, the Ontology based SMS Controller validates the sender number with the spam blacklist numbers, if the number is found in the blacklist numbers, the SMS is send to SPAM folder without further processing. The Ontology based SMS Controller also provides user the option to SPAM all SMS messages from unknown numbers or specific numbers or weird numbers. If this option is selected by the user, all SMS belonging to these categories will be send to SPAM folder without further processing.

In Step 2, all standard stop-list / stemmer words like ("is", "the", "on", "and", "in", "with", "for", "by"…) are eliminated from the SMS Text. In Step 3, homogeneous words like {("chat", "chatting", "chatted"), ("Advertize", "Advertizing", "Advertized")} are all substituted by the single word "chat" and "Advertize" respectively. Also, multiple entries for each word are eliminated from the SMS text.

- **Content Analysis:** This module uses the filtered SMS text from the previous step which contains n keywords where each keyword can express *n* possible meanings. In order to assign proper meaning to each keyword, every keyword is compared with every other keyword and most related sense (i.e. semantically related) is selected. To calculate the most related sense the shortest path (i.e. minimum number of nodes present in the path connecting the keywords is used); WordNet [18] is used for this purpose. In the next step, Concept set is generated which contains either the original keywords or Lowest Super Ordinate (LSO) for each pair of keywords, the selection depends on the parameter *h,* for more details please see [19].

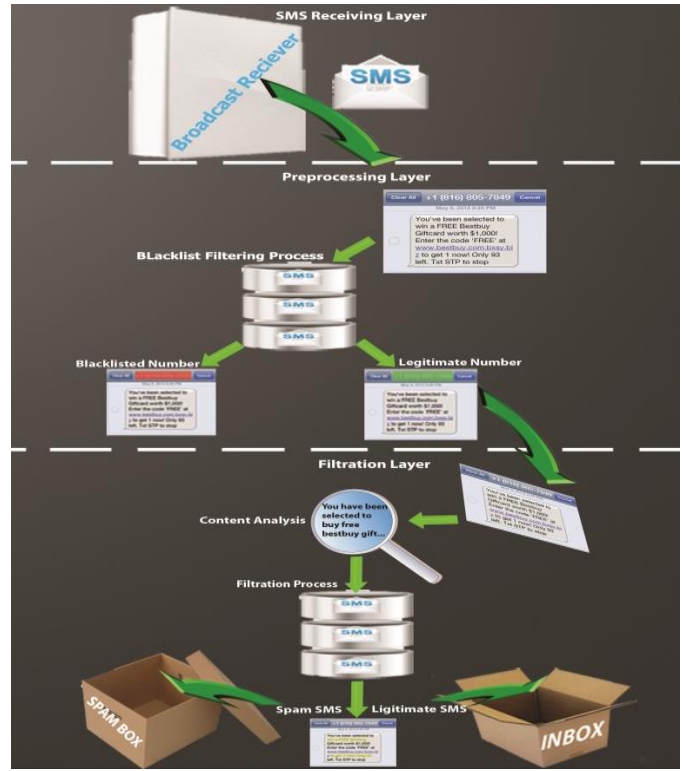

Fig. 2. Ontology based SMS Controller Spam Module

Spam concepts which include spam keywords, their synonym and hypernym are loaded from the ontology and stored in spam set. Each concept is compared with the spam concepts one by one and matches are stored in a separate resultant set with labels O (Original Keyword), S (Keyword Synonym) or H (Keyword Hypernym). The labels are assigned on the basis of comparison; if the concept is matched with spam keyword, O is assigned. Similarly if the concept is match with Spam keyword synonym or hypernym, S or H is assigned respectively. Each concept in the resultant set is assigned a score based on the assigned label (i.e. O=1, S=0.50 and H=0.25). The collective spam score of the resultant set is calculated by adding the all individual concept scores.

- **Classification and Ontology Enhancement:** This module uses the resultant set extracted from the previous step, to classify the SMS as spam or not. Furthermore, if the SMS is classified as spam, the ontology will be enhanced by adding new concepts (i.e. Original keywords, synonym and hypernym). In order to classify the SMS as spam, the Collective Spam score (CS) calculated in the previous set is used, if condition (where is configurable) is satisfied, the SMS is classified as Spam and forwarded to Spam folder. Once the SMS is classified as Spam, Keywords Synonyms and Hypernyms are extracted from WordNet and new concepts (i.e. Keywords, Synonyms and Hypernyms) are added to the ontology knowledge base.

*B. Group Chat*

The prerequisite of using this feature is that all participating users should have Ontology based SMS Controller installed on the smart phone. One of the application user has to start the Group chat by sending "join group chat" invitation to the others. Invitation is sent through SMS message and for this purpose a special SMS is send to the invitee which Ontology based SMS Controller interprets and asks the user to join the group chat. The user can accept or reject the request, if the user accepts, the details of new user is send to all the active members of the group chat and chat window is loaded on the new user's Smartphone.

Similarly, if any active user during the group chat closes the application, the details of disconnected user is send to all members of the group chat. Each group chat is assigned unique chat code to the same and each SMS message send / received from chat window contains this unique chat code, which makes it easy to identify; to which chat this message belongs. Each member sets a nick at the start of chat, and these are displayed on the chat window instead of the numbers.

Each incoming SMS is interrupted by Ontology based SMS Controller and it validates the type of the SMS message (i.e. Invitation, Chat Message, or Normal Message) and perform actions accordingly. If the SMS is chat message, it forwards the same to the corresponding chat window and delete it from the inbox. If the SMS is invitation SMS, it displays the invitation to the user and wait for the response. Based on user response it either opens the chat windows or sends rejection message. When a user joins or leaves the chat, all other members are informed and the list of chat members is updated accordingly.

*C. Auto Reply*

Ontology based SMS Controller sends auto-reply according to the user-defined profiles; the user is responsible to create auto-reply groups and reply messages for each group. If the auto-reply mode is on and no auto-reply profile is activated, a default auto-reply message is sent otherwise the user defined auto-reply message according to the profile is send. User can create groups and add numbers in these groups from contact list.

*D. Event-Based Messages*

Ontology based SMS Controller automatically synchronizes itself with the calendar and generates automatic SMS based on Events. User is responsible to define event(s) by setting date / time of the event(s) and the message to be send. User can add group(s) to an event by selecting from list of available groups. Message defined against the event is send to all members of the group(s) associated with the event. Birthday event is predefined in the application, which picks the birthdays of the contacts (if available) and create birthday event for each of them. User can define the birthday message for the birthday event, if no message is defined; default birthday wish is send automatically on respective birthdays.

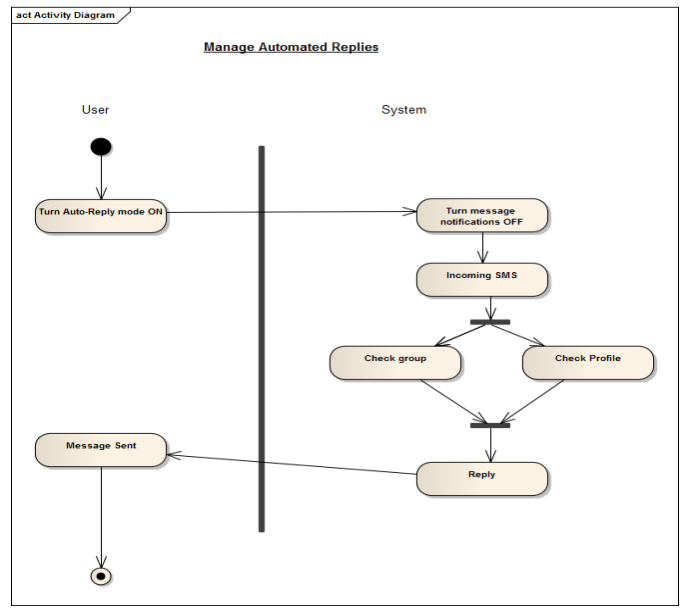

Fig. 4. Auto-Reply flow

IV. SIMULATION AND EXPERIMENTAL RESULTS

The Application is developed on Android version 4.1 (Jelly Beans) and is compatible with all the next versions of Android (min SDK 8). Android Virtual Device (AVD) manager is installed with Android Software Development kit which allows the programmer to create an AVD for specific version of Android. The simulator used for testing and debugging of the proposed application is AVD 4.1.

Initially for the experimentation, we built the ontology concepts using one hundred known spam messages; afterward we tested the proposed solution on large number of SMS, figure 5 shows spam detection percentage over number of SMS, as shown in figure 5 the proposed solution spam detection percentage over number of SMS increases as ontology knowledgebase is enhanced (i.e. new spam concepts are updated in the ontology).

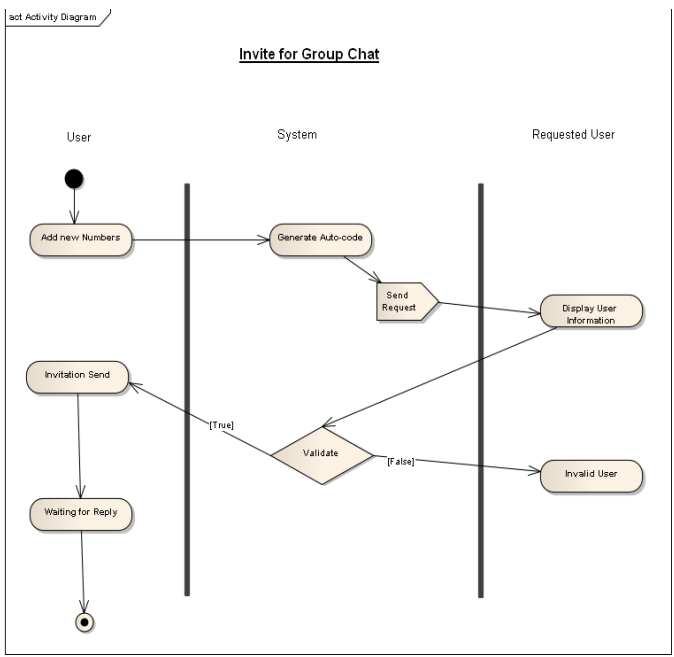

Fig. 3. Group Chat Flow

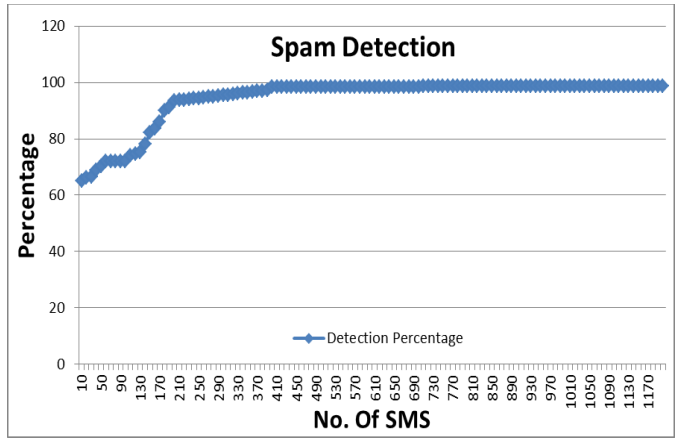

Fig. 5. Spam Detection vs No. of SMS

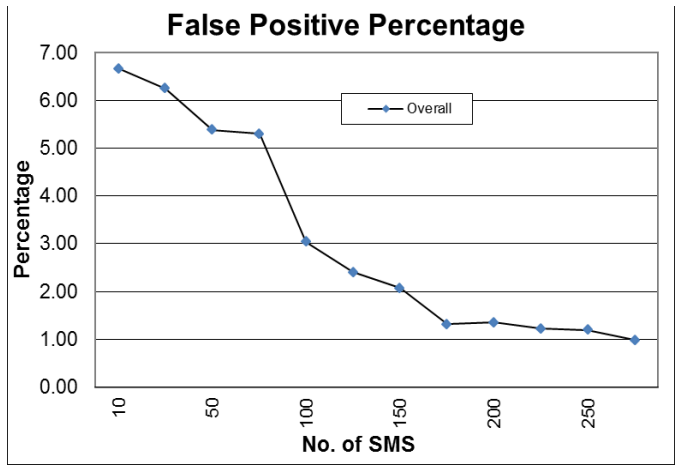

Fig. 6. False Positive Percentage vs No. of SMS

Figure 6 shows false positive percentage over number of SMS, as shown in figure 6 the proposed solution false positive percentage over number of SMS decreases as the system receives feedback from user over wrongly predicted SMS(s) which helps in updating the ontology knowledgebase by removing concepts related to wrongly predicted SMS(s).

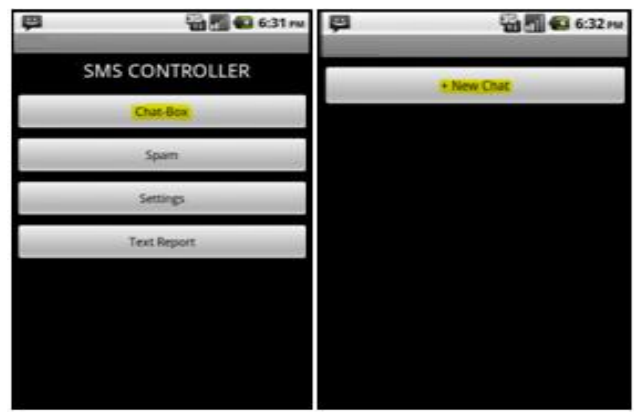

Fig. 7. (a) Main View (b) New Chat Window

Figure 7(a) shows the main view of the application, if user press Chat-Box new window opens as shown in figure 7(b). If user press New Chat button, new chat is started.

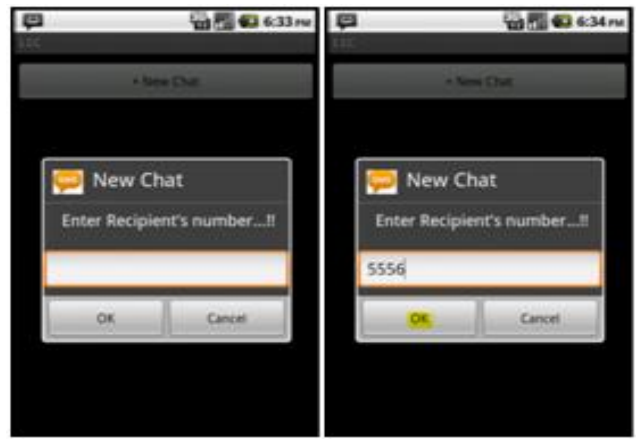

Fig. 8. New Chat

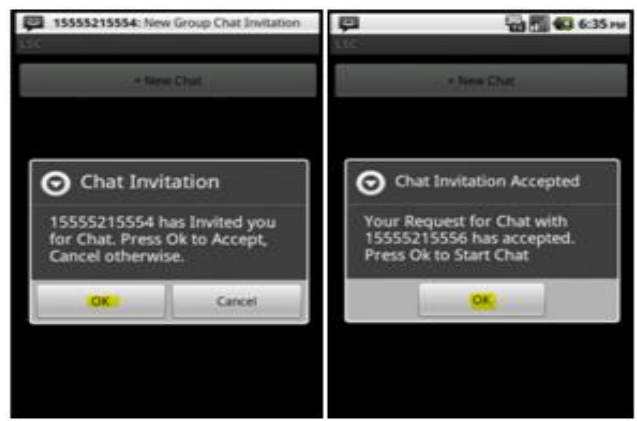

Fig. 9. (a) Chat Invitation (b) Invitation Accepted

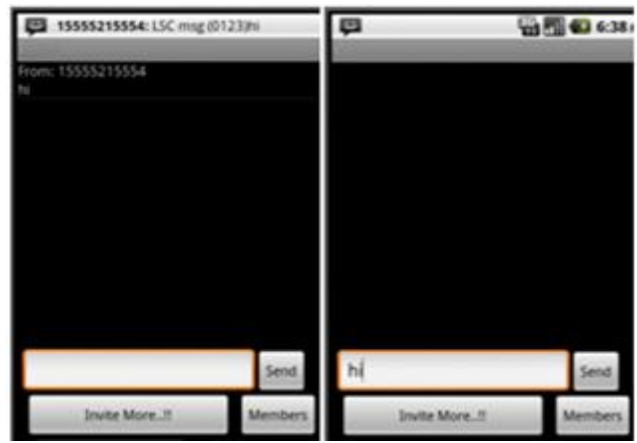

Fig. 10. Group Chat Window

Figure 8 shows new chat window where the user needs to enter the Recipient's number and press ok to invite the same to group chat.

Figure 9(a) shows the invitation for group chat message which appears on the recipient's device, if the user accepts the invitation figure 9(b) is shown on the sender device. After handshaking the group chat is started and chat window appears on each participating device (i.e. Sender & Receivers) as shown in figure 10. User can write the text in text Box and use the send button to send the message. The message is send to all recipients by using normal SMS and is shown on each participating device. New members can be added at any time using the invite more feature.

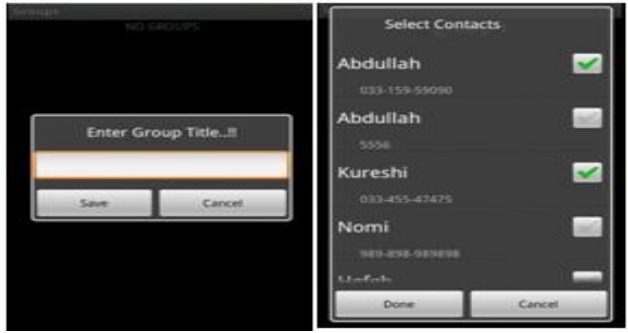

Fig. 11. (a) Group title (b) Group contacts

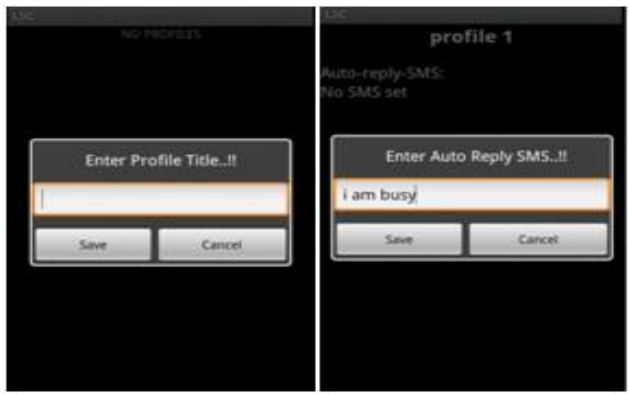

Fig. 12. (a) Profile title (b) Auto-reply SMS

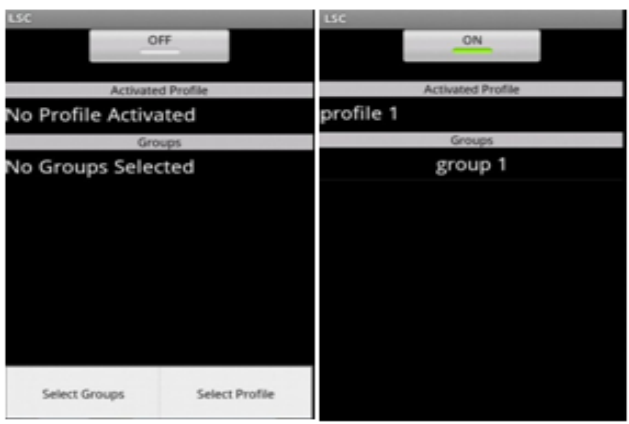

Fig. 13. Auto-reply

The user can create new groups, just he needs to give the title of the group and add contacts in the group as shown in figure 11(a) and (b) respectively. Contacts can be added to group by going to the desired group and selecting "add contact" from menu. Similarly user can create new profile(s), each profile contains profile title and auto-reply SMS. The user needs to enter the title of the profile and auto-reply SMS as shown in figure 12(a) and (b) respectively. Auto-Reply SMS can be added by going to the desired profile and selecting "Select auto-reply SMS" from menu.

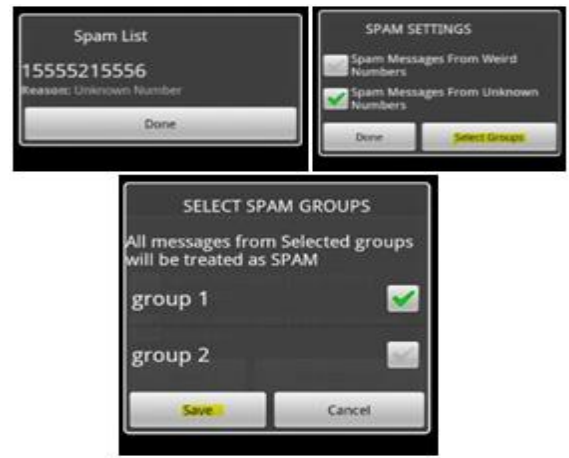

Fig. 14. Spam Options

Auto-reply mode can be activated by clicking the toggle (on/off) button in figure 13. Left window of figure 13 shows the view where auto-reply mode is off, no group and profile selected. Right window of figure 13 is showing the view where mode is on, one group is selected and profile is activated.

Spam Filter can be turned on by clicking toggle button on the spam window. Spam folder contains all the spam messages. Spam List contains all the numbers which are marked as spam. Spam Settings allows user to spam messages from weird numbers or unknown numbers by checking the checkboxes. A group can be added and marked as spam as well as shown in figure 14.

Similarly new events can be added by the user specifying event title, date / time, profile / SMS and group(s) / numbers associated with the event. The application sends the specified SMS to all the members associated with the event on the event date / time.

V. CONCLUSION

With the evolution in Smartphone era, leading IT companies and researchers have proposed many efficient applications; one of them is SMS Manager which helps to manage the SMS on smart phones. In this paper, we proposed Ontology based SMS Controller which is all in one SMS manager and includes new features like content based SMS Detection, Group chat etc. SMS Spam classification algorithm of Ontology based SMS Controller analyses the text of SMS and uses ontology to classify it as Spam or legitimate. The proposed algorithm has been tested on large number of test cases; the experimental results are satisfactory and supports the implementation of the solution.

REFERENCES

[1] Yung Fu Chang, C.S. Chen, Hao Zhou, "Smart phone for mobile commerce", Computer Standards & Interfaces, Volume 31, Issue 4, June 2009, Pages 740-747.

[2] Yung-Fu Chang, C.S. Chen, "Smart phone – the choice of client platform for mobile commerce", Computer Standards & Interfaces, Volume 27, Issue 4, April 2005, Pages 329-336.

[3] Lorena Otero-Cerdeira, Francisco J. Rodríguez-Martínez, Alma Gómez-Rodríguez, "Ontology matching: A literature review", Expert Systems with Applications, Volume 42, Issue 2, 1 February 2015, Pages 949-971.

[4] Carla Faria, Ivo Serra, Rosario Girardi, "A domain-independent process for automatic ontology population from text", Science of Computer Programming, Volume 95, Part 1, 1 December 2014, Pages 26-43.

[5] Umar Manzoor, Samia Nefti, Yacine Rezgui "Categorization of malicious behaviors using ontology-based cognitive agents", Data & Knowledge Engineering, Volume 85, May 2013, Pages 40-56.

[6] Francesco Rea, Samia Nefti-Meziani, Umar Manzoor, Steve Davis "Ontology enhancing process for a situated and curiosity-driven robot", Robotics and Autonomous Systems, Volume 62, Issue 12, December 2014, Pages 1837-1847.

[7] Mohamed Yehia Dahab, Hesham A. Hassan, Ahmed Rafea "TextOntoEx: Automatic ontology construction from natural English text" Expert Systems with Applications, Volume 34, Issue 2, February 2008, Pages 1474-1480.

[8] Ming-Hsiung Hsiao, Liang-Chun Chen "Smart phone demand: An empirical study on the relationships between phone handset, Internet access and mobile services" Telematics and Informatics, Volume 32, Issue 1, February 2015, Pages 158-168.

[9] Apple Inc, (2012), http://www.apple.com/ (Access Date: 12-06-2014)

[10] Apple iPhone, (2012), http://www.apple.com/iphone/ (Access Date: 12-06-2014)

[11] Short Messaging Service (SMS), 2012, http://en.wikipedia.org/wiki/SMS (Access Date: 02-02-2014)

[12] Droid Mate, "Anti Spam SMS and Private Box", https://market.android.com/details?id=org.baole.app.antismsspam&hl=en (Access Date: 15-01-2014)

[13] Handcent Market - Handcent SMS, https://market.android.com/details?id=com.handcent.nextsms&hl=en (Access Date: 15-01-2014)

[14] Kirill kruchinkin, "SMS AutoReply", http://www.appbrain.com/app/sms-autoreply/auto.msg (Access Date: 15-01-2014)

[15] John Tsau, "Intelligent AutoReply", https://market.android.com/details?id=com.jtsau.autoReply, (Access Date: 16-01-2014)

[16] GO Dev Team - GO SMS PRO, https://market.android.com/details?id=com.jb.gosms&hl=en, (Access Date: 15-01-2014)

[17] Inna Novalija, Dunja Mladenić, Luka Bradeško "OntoPlus: Text-driven ontology extension using ontology content, structure and co-occurrence information", Knowledge-Based Systems, Volume 24, Issue 8, December 2011, Pages 1261-1276.

[18] WordNet (2014), http://wordnet.princeton.edu/

[19] Samia Nefti, M. Oussalah, Yacine Rezgui "A modified fuzzy clustering for documents retrieval: application to document categorization", Journal of the Operational Research Society, Volume 60, Number 3, pp. 384-394, March 2009.

[20] Manzoor, U.; Khan, M.; Qureshi, A.; ul Ghani, N., "Luxus SMS controller for android based smart phones," International Conference on Information Society (i-Society), pp. 315-320, 25-28 June 2012.

[21] Kwanho Kim, Beom-suk Chung, Yerim Choi, Seungjun Lee, Jae-Yoon Jung, Jonghun Park "Language independent semantic kernels for short-text classification", Expert Systems with Applications, Volume 41, Issue 2, 1 February 2014, Pages 735-743.

[22] Duc-Thuan Vo, Cheol-Young Ock "Learning to classify short text from scientific documents using topic models with various types of knowledge", Expert Systems with Applications, Volume 42, Issue 3, 15 February 2015, Pages 1684-1698.

[23] Umar Manzoor, Samia Nefti, Yacine Rezgui "Autonomous Malicious Activity Inspector – AMAI" Natural Language Processing and Information Systems, Lecture Notes in Computer Science Volume 6177, 2010, pp 204-215.

[24] Umar Manzoor, Samia Nefti "iDetect: Content Based Monitoring of Complex Networks using Mobile Agents", Applied Soft Computing, Volume 12, Issue 5, May 2012, Pages 1607–1619.

[25] Hao-jin TANG, Dan-feng YAN, Yuan TIAN "Semantic dictionary based method for short text classification", The Journal of China Universities of Posts and Telecommunications, Volume 20, Supplement 1, August 2013, Pages 15-19.

[26] Umar Manzoor, Samia Nefti "Autonomous agents: Smart network installer and tester (SNIT)", Expert Systems with Applications, Volume 38, Issue 1, January 2011, Pages 884–893.

[27] Umar Manzoor, Bassam Zafar "Multi-Agent Modeling Toolkit – MAMT", Simulation Modelling Practice and Theory, Volume 49, December 2014, Pages 215–227.

[28] Lili Yang, Chunping Li, Qiang Ding, Li Li "Combining Lexical and Semantic Features for Short Text Classification", Procedia Computer Science, Volume 22, 2013, Pages 78-86.